\documentclass{article}

\usepackage[preprint,nonatbib]{neurips_2020}

\usepackage[utf8]{inputenc} 
\usepackage[T1]{fontenc}    
\usepackage{hyperref}       
\usepackage{url}            
\usepackage{booktabs}       
\usepackage{amsfonts}       
\usepackage{nicefrac}       
\usepackage{microtype}      

\usepackage{graphicx}
\usepackage{enumitem}
\usepackage{algorithmic}
\usepackage{algorithm}
\usepackage{amsmath}
\usepackage{booktabs} 
\usepackage{multirow}
\usepackage[lofdepth,lotdepth]{subfig}
\usepackage{comment}
\newcommand{\tablestyle}[2]{\setlength{\tabcolsep}{#1}\renewcommand{\arraystretch}{#2}\centering\scriptsize}
\usepackage{xcolor}

\newcommand{\etal}{\textit{et al}.~}

\newcommand{\ieno}{\textit{i}.\textit{e}.}

\newcommand{\egno}{\textit{e}.\textit{g}.} 

\title{Feature Alignment and Restoration for Domain Generalization and Adaptation}

\author{{Xin Jin{$^{1,2}$}\thanks{This work was done when Xin Jin was an intern at MSRA.}} \qquad Cuiling Lan{$^{2}$}\thanks{Corresponding Author.} \qquad   Wenjun Zeng{$^{2}$} \qquad  Zhibo Chen{$^{1{\dagger}}$} \\
$^{1}$University of Science and Technology of China\and
$^{2}$Microsoft Research Asia, Beijing, China\\
\texttt{jinxustc@mail.ustc.edu.cn\quad \{culan,wezeng\}@microsoft.com\quad chenzhibo@ustc.edu.cn}
}

\begin{document}

\maketitle

\begin{abstract}


For domain generalization (DG) and unsupervised domain adaptation (UDA), cross domain feature alignment has been widely explored to pull the feature distributions of different domains in order to learn domain-invariant representations. However, the feature alignment is in general task-ignorant and could result in degradation of the discrimination power of the feature representation and thus hinders the high performance. In this paper, we propose a unified framework termed Feature Alignment and Restoration (FAR) to simultaneously ensure high generalization and discrimination power of the networks for effective DG and UDA. Specifically, we perform feature alignment (FA) across domains by aligning the moments of the distributions of attentively selected features to reduce their discrepancy. To ensure high discrimination, we propose a Feature Restoration (FR) operation to distill task-relevant features from the residual information and use them to compensate for the aligned features. For better disentanglement, we enforce a dual ranking entropy loss constraint in the FR step to encourage the separation of task-relevant and task-irrelevant features. Extensive experiments on multiple classification benchmarks demonstrate the high performance and strong generalization of our FAR framework for both domain generalization and unsupervised domain adaptation.

\end{abstract}

\section{Introduction}

Deep neural networks (DNNs) have advanced the state-of-the-art for a wide variety of machine-learning problems and applications. The trained models typically perform well when the test data follows a similar distribution as the training data \cite{krizhevsky2012imagenet}, but suffer from significant performance degradation (poor generalization capability) when testing on a previously unseen data distribution \cite{long2016unsupervised}. The data may be captured in different environments or obtained by different manners (\egno, photo, sketch, painting), which exhibit domain shifts. To address such domain shift problem, two related fields have been studied intensively: domain generalization (DG) \cite{muandet2013domain,li2017deeper,shankar2018generalizing,li2018learning,carlucci2019domain,li2019episodic} and unsupervised domain adaptation (UDA) \cite{pan2009survey,ganin2014unsupervised,long2015learning,long2016unsupervised,saito2018maximum,hoffman2018cycada,peng2019moment,xu2019larger,wang2019transferable}. DG and UDA both aim to bridge the source and target domains by learning domain-invariant feature representations, where DG can only exploit source domain data while UDA can also exploit the unlabeled training data of the target domain for training/fine-tuning.

Feature regularization based methods have been widely investigated to mitigate the domain gap. Several methods incorporate the distribution distance metric, such as Maximum Mean Discrepancy (MMD), into model as loss to diminish the domain discrepancy for DG and UDA \cite{tzeng2014deep,long2017deep}. Some other methods introduce different learning schemes to align the source and target domains, including aligning second order correlation \cite{sun2016return,sun2016deep,peng2018synthetic}, moment matching \cite{zellinger2017central,peng2019moment}, adversarial domain confusion \cite{ganin2014unsupervised,ganin2016domain,tzeng2017adversarial,saito2018maximum}, and GAN-based alignment \cite{zhu2017unpaired,liu2017unsupervised,hoffman2018cycada}. They intend to drive the DNNs
to learn domain-invariant features for improving its generalization and transferability \cite{hoffman2013efficient,baktashmotlagh2013unsupervised,li2018domain}.

\begin{figure}
  \centerline{\includegraphics[width=1.0\linewidth]{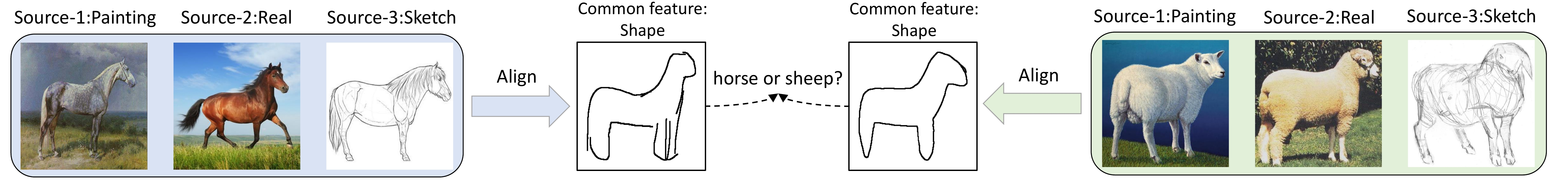}}
    \caption{Motivation: only depending on the aligned feature seems not enough for clear classification.}
\label{fig:motivation}
\vspace{-4mm}
\end{figure}

However, the aligning in general overlooks a side effect, \ieno, the loss of task-relevant discriminative information. The data of different domains has some domain-shared characteristics and also some domain-specific characteristics. Aligning the features is to obtain a shared sub-space for all these domains, where the domain-specific information tends to be excluded to assure the alignment. Aligning the features across domains through some constraints drives the features to lie in a shared space. However, since the purpose of alignment is not task-specific, it could lose some information, including task-relevant information. As illustrated by the toy example in Figure \ref{fig:motivation}, the common space after domain alignment of the three source domains (including painting, real, sketch) could be that related with \emph{shape}. However, besides the \emph{shape}, the texture information is also discriminative to distinguish between the `horse' and `sheep' for the domain of real and painting. Among the previous works, the restoration of discriminative information is overlooked even though important.

In this paper, we propose a novel unified framework termed Feature Alignment and Restoration (FAR) to perform feature aligning and restoration to simultaneously enhance cross domain generalization and preserve feature discrimination for effective DG and UDA classification. Figure \ref{fig:pipeline} illustrates the pipeline. It consists of four parts in order: feature extraction, feature alignment (FA), feature restoration (FR), and classification. In the FA phase, similar to \cite{peng2019moment}, we align the feature distributions across different domains by matching the moments of the multiple distributions. Different from \cite{peng2019moment}, we enforce the alignment on the adaptively selected features rather than on the original features in order to reserve the comprehensive original information and find a subspace (selected features) that is suitable for alignment. Since the alignment is task-ignorant, the aligned features lose some discriminative information. We propose a Feature Restoration (FR) to further distill task-relevant feature from the residual to compensate for the aligned feature. Moreover, within FR, for better disentanglement, we enforce a dual ranking entropy loss constraint to encourage the separation of task-relevant and task-irrelevant features. Finally, a domain-shared classifier is used to classify the restored features, trained with the guidance of an expert of each domain. We validate the effectiveness of the FAR framework on multiple widely-used benchmarks for both the DG and UDA settings. FAR significantly outperforms the state-of-the-art DG and UDA approaches. We summarize our main contributions as follows:

\begin{itemize}[leftmargin=*,noitemsep,nolistsep]

\item We introduce the concept of feature alignment and restoration (FAR) to the cross-domain classification task for effective DG or UDA, by ensuring high generalization and discrimination power of the learned feature within a unified framework.

\item In the FR phase, we introduce a dual ranking entropy loss constraint to facilitate the restoration of the task-relevant information and better disentanglement.

\item To keep the simplicity, as a minor insight, we avoid the use of multiple classifiers for the inference by training a shared classifier with the guidance of the expert teacher of each domain. 

\end{itemize}


\section{Related Work}




\noindent\textbf{Unsupervised Domain Adaptation (UDA).}
Unsupervised domain adaptation is a target domain annotation-free transfer learning task, where there are domain shifts/gaps between the training source dataset and the testing target dataset. Motivated by the seminal theory work \cite{ben2010theory}, many advanced UDA methods seek to learn domain-invariant features by reducing the distribution discrepancy between source and target features. Some methods minimize distribution divergence with the losses of maximum mean discrepancy (MMD) \cite{long2015learning,long2016unsupervised,long2017deep}, second order correlation CORAL \cite{sun2016return,sun2016deep,peng2018synthetic}. Some others learn a domain discriminator in an adversarial manner to achieve domain confusion \cite{ganin2014unsupervised,tzeng2017adversarial,zhang2018collaborative,zhao2018adversarial}. For better feature alignment, Maximum Classifier Discrepancy (MCD) \cite{saito2018maximum} maximizes the discrepancy between two classifiers while minimizing it with respect to the feature extractor. Similarly, Minimax Entropy (MME) \cite{saito2019semi} maximizes the conditional entropy on unlabeled target data with respect to the classifier and minimizes it with respect to the feature encoder. M3SDA \cite{peng2019moment} is one of the latest multi-source UDA methods and it minimizes the moment distance among the source and target domains and per-domain classifier is used and optimized as in MCD to enhance alignment.
%


\noindent\textbf{Domain Generalization (DG).} DG considers a more challenging yet attractive setting where the target data is unavailable during training. Many DG methods share the similar ideas of distribution alignment with UDA to learn domain-invariant features by minimizing distances among the multiple source domains \cite{muandet2013domain,ghifary2016scatter,motiian2017unified,shankar2018generalizing}. Some other DG methods attempt to explore optimization/training strategies, \egno, through meta-learning \cite{li2018learning}, episodic training \cite{li2019episodic}, and adaptive ensemble learning \cite{zhou2020domain}.

For UDA or DG classification, these alignment-based methods typically consider to use feature regularizations to achieve domain aligning but inevitably lose some discriminative information. How to fully exploit all the discriminative information in both aligned and un-aligned features remains under-explored. Jin \etal propose a domain generalization method for person re-identification \cite{jin2020style}, which reduces the style discrepancy by instance normalization and restitutes the identity-relevant features. Different from it, we address the more general domain gaps in classification and propose a unified framework with feature alignment and restoration to support both the domain generalization and adaptation tasks. The general classification is more challenging than re-identification where the latter is a fine-granularity retrieval task and the former needs to handle larger intra-class variations.

\noindent\textbf{Feature Disentanglement.} DNNs are known to extract features where multiple hidden factors are highly entangled \cite{mathieu2016disentangling}. Some works \cite{makhzani2015adversarial,odena2017conditional,liu2018unified,lee2018diverse} explore the learning of interpretable representations, by leveraging generative adversarial networks (GANs) \cite{goodfellow2014generative} or variational autoencoders (VAEs) \cite{kingma2013auto}. 
In this paper, different from prior works, we aim to disentangle the task-relevant features from the residual features (the difference between the original features and the aligned features) and return them for effective classification. We propose a dual ranking entropy loss constraint in the feature restoration phase, by enforcing higher discrimination of the feature after the restoration than before. 

\section{Proposed Unified Feature Alignment and Restoration Framework}


\noindent\textbf{Problem definition.}
Given $K$ source domain $\mathcal{D}_s = \{\mathcal{D}_{s1}, \cdots, \mathcal{D}_{s_K} \}$ of labeled images and a target domain ${{\cal D}_t}$ of unlabeled images. Due to the domain shifts/gaps, models trained in source domains in general suffer from performance degradation when tested on a target domain. The goal of this work is to design a unified framework for both domain generalization (DG) and unsupervised domain adaptation (UDA) classification. 

\begin{figure}[t]
  \centerline{\includegraphics[width=1.0\linewidth]{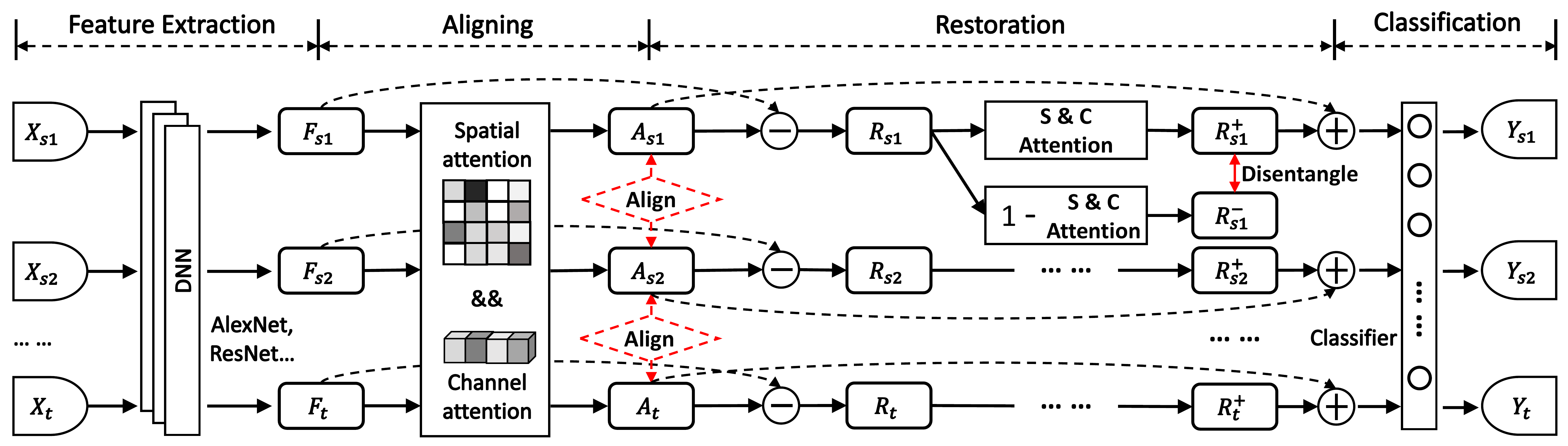}}
    \caption{Flowchart of the Feature Alignment and Restoration (FAR) framework.}
\label{fig:pipeline}
\vspace{-5mm}
\end{figure}

\noindent\textbf{Overview.} 
Figure \ref{fig:pipeline} shows the pipeline of our feature alignment and restoration (FAR) framework when applied to the classification task. We split the classification pipeline into four parts: feature extraction, feature alignment (FA), feature restoration (FR), and classification. In particular, FA and FR aim to boost the generalization and discrimination capability of the network. In the FA phase, we first leverage spatial and channel attention to adaptively select features from the extracted feature map $F$ of the backbone network and enforce across domain alignment constraints on them. Then, the FR step distills task-relevant (discriminative) features from the residual to compensate for the aligned features. Moreover, in the FR phase, we design a dual ranking entropy loss constraint to promote the distillation of the task-relevant features from the residual. Finally, a domain-shared classifier is used to classify the restored features, trained with the guidance of an expert of each domain. 

\subsection{Feature Alignment and Restoration}\label{subsec:FAR}

In the fields of domain generalization and adaptation, some theoretical analyses~\cite{ben2010theory,Mansour_nips2018,crammer2008learning,Zhao2017MultipleSD, NIPS2018_8046} show that reducing the dissimilarity among domains improves the generalization ability. Ben-David \etal~\cite{ben2010theory} pioneer this direction by introducing an $\mathcal{H}\Delta\mathcal{H}$-divergence among the source domains and target domain. The recent work~\cite{peng2019moment} extends the prior theoretical analysis to the case of moment-based divergence between source and target domains, which matches the moments of the distributions across domains to achieve feature alignment. However, these works focus only on aligning domains. They overlook the side effect of the loss of task-relevant information due to the alignment.

To address this problem, we propose to restore the discriminative features from the discarded domain-specific information, by disentangling it into task-relevant features and task-irrelevant features with a dual ranking entropy loss constraint. 



\noindent\textbf{Feature Alignment (FA) Phase.} 
FA aims to reduce domain discrepancy and get rid of domain-specific information to makes feature domain-invariant. As shown in Figure \ref{fig:pipeline}, for the FA phase, we denote the input feature map (obtained by the Feature Extraction step from the backbone network) by $F \in \mathbb{R}^{h\times w \times c}$ with width $w$, height $h$, and the number of channels $c$. We denote the aligned feature map by $A \in \mathbb{R}^{h\times w \times c}$. We use different subscripts to represent the domain of data, \egno, $F_{s1}$ denotes a feature map extracted from data of the first source domain.

To allow the feature $F$ to preserve comprehensive information, instead of on $F$, we enforce the alignment constraints on the adaptively selected feature $A$, which is obtained by gating the comprehensive feature $F$ with the tool of spatial and temporal attention in parallel \cite{hu2018squeeze,woo2018cbam,fu2019dual,liu2019adaptive} (see Supplementary about the formulation). Such attentive gating facilitates the adaptive selection of feature sub-spaces that are suitable for alignment.  





On the attentively selected feature $A$, similar to \cite{peng2019moment}, we add the  constraints $\mathcal{L}_{align}$ which align the moments of the feature distributions across different domains (see Supplementary for more details).

\noindent\textbf{Feature Restoration (FR) Phase.}
The feature alignment constraints are task-ignorant, this inevitably discards some task-relevant information. We propose to compensate the network with such task-relevant feature by distilling it from the residual features $R$, which is defined as the difference between the comprehensive original feature $F$ and the aligned feature $A$ as $R =F-A$.

Given the residual $R$, we further disentangle it into task-relevant feature $R^+ \in \mathbb{R}^{h\times w\times c}$ and task-irrelevant feature $R^- \in \mathbb{R}^{h\times w\times c}$. To enable the content-adaptive disentanglement, we use a  spatial and channel attention module (similar to the attention in the FA, see Supplementary) as the gate to obtain task-relevant feature $R^+$ and the remaining task-irrelevant feature $R^-$ as
\begin{equation}
    \begin{aligned}
        R^+ =  Gate(R) \cdot R, \hspace{0.5cm}
        R^- =  (1 - Gate(R)) \cdot R, 
    \end{aligned}
    \label{eq:seperation}
\end{equation}
where $Gate(R)$ represents the spatial and channel attention response on $R$. By adding the disentangled task-relevant feature $R^+$ to the aligned feature $A$, we obtain the restoration feature $A + R^+$.

\noindent\textbf{Dual Ranking Entropy Loss Constraint.}
In order to facilitate the disentanglement into task-relevant and task-irrelevant features, we design a dual ranking entropy loss constraint.

Intuitively, after adding the task-relevant feature $R^+$ to the aligned feature $A$, we expect the \emph{enhanced} feature becomes more discriminative and thus the entropy of predicted class likelihood would become smaller, with less ambiguity. On the other hand, after adding the task-irrelevant feature (interference) $R^-$ to the aligned feature $A$, the \emph{contaminated} feature becomes less discriminative and thus the entropy would become larger. Therefore, we design a dual ranking entropy loss $\mathcal{L}_{DRE} = \mathcal{L}_{DRE}^+ + \mathcal{L}_{DRE}^-$.

We pass the spatially average pooled \emph{enhanced} feature vector $\mathbf{f^+} = pool(A+R^+) \in \mathbb{R}^c$ through the final classifier $C$ and calculate its entropy. We denote the entropy functions as $E(\cdot)=-p(\cdot) \log p(\cdot)$. Similarly, the \emph{contaminated} feature vector can be obtained by $\mathbf{f^-} = pool(A+R^-)$. The reference feature vector $\mathbf{f} = pool(A)$. $\mathcal{L}_{DRE}^+$ and $\mathcal{L}_{DRE}^-$ are defined as:
\begin{equation}
    \begin{aligned}
    \mathcal{L}_{DRE}^+ &= Softplus( E(C(\mathbf{f^+})) - E(C(\mathbf{f})), \hspace{0.2cm} \mathcal{L}_{DRE}^- = Softplus(E(C(\mathbf{f})) - E(C(\mathbf{f^-}))).
    \end{aligned}
    \label{eq:seperation}
\end{equation}
Here $Softplus(\cdot) = ln(1+exp(\cdot))$ is a monotonically increasing function that aims to reduce the optimization difficulty by avoiding negative loss values.

\subsection{Training} 
Our FAR framework is trained in an end-to-end fashion. For DG, the total loss $\mathcal{L}$ is a weighted summation of the alignment loss $\mathcal{L}_{align}$, dual ranking entropy loss $\mathcal{L}_{DRE}$ and the basic classification loss $\mathcal{L}_{cls}$. The training data is all from the source domains. For UDA, as shown in Figure \ref{fig:pipeline}, the target domain data without labels is also used and the loss for such data consists of $\mathcal{L}_{align}$ and $\mathcal{L}_{DRE}$. Please see our Supplementary for the weight determination on each loss and pseudo code.



Note that the network parameters are all shared for different domains. To enable fast convergence and learning of the knowledge from the feature $F$ which contains comprehensive information (both domain invariant and domain specific), as a design choice, we assign a classifier for each source domain on the feature $F$ and let them teach the final shared classifier by enforcing L1-distance consistency $\mathcal{L}_{consist}$ on the predicted probability from the classifiers. In the inference phase, the classifiers for individual source domains are not needed.

\section{Experiments}

\begin{figure}
  \centerline{\includegraphics[width=1.0\linewidth]{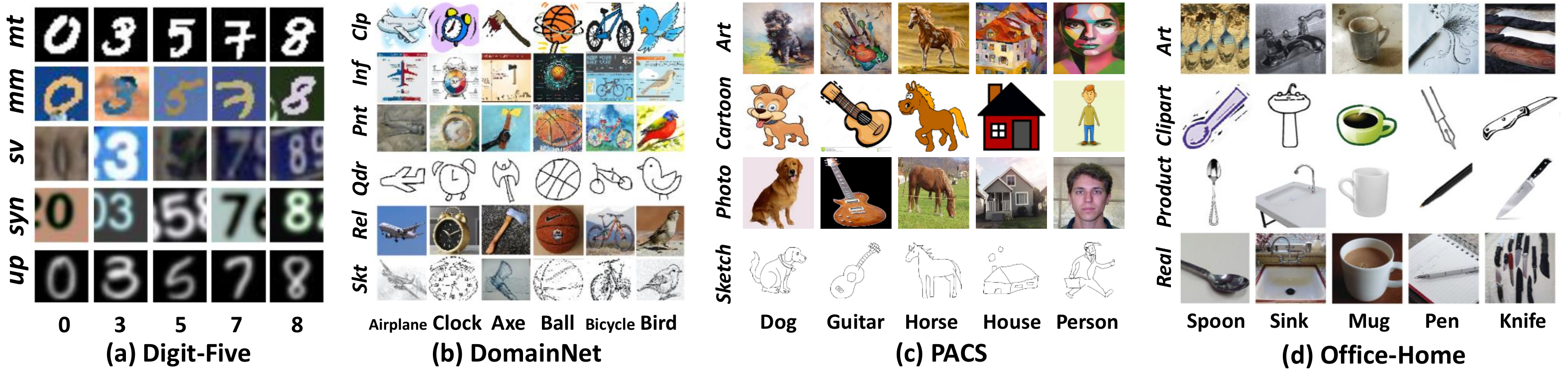}}
    \caption{Four classification datasets (first two for UDA and last two for DG). (a) Digit-Five, includes MNIST \cite{lecun1998mnist} (\textit{\textbf{mt}}), MNIST-M \cite{ganin2015unsupervised} (\textit{\textbf{mm}}), USPS \cite{hull1994database} (\textit{\textbf{up}}), SVHN \cite{netzer2011svhn} (\textit{\textbf{sv}}), and Synthetic \cite{ganin2015unsupervised} (\textit{\textbf{syn}}). (b) DomainNet, includes {Clipart} (\textit{\textbf{clp}}), {Infograph} (\textit{\textbf{inf}}), {Painting} (\textit{\textbf{pnt}}), {Quickdraw} (\textit{\textbf{qdr}}), {Real} (\textit{\textbf{rel}}), and {Sktech} (\textit{\textbf{skt}}). Considering the required huge computation resources, we use a subset of DomainNet (\ieno, mini-DomainNet) following \cite{zhou2020domain} for experiments. (c) PACS, includes \textit{Sketch}, \textit{Photo}, \textit{Cartoon}, and \textit{Art}. (d) Office-Home, includes {Real-world} (\textit{Real}), \textit{Product}, \textit{Clipart}, and \textit{Art}.}
\label{fig:datasets}
\vspace{-4mm}
\end{figure}

We describe the datasets and implementation details in Sec. \ref{subsec:dataset}. We validate the effectiveness and superiority of our FAR method in Sec. \ref{subsec:ablation}. We study some design choices in Sec. \ref{subsec:design}. Moreover, Sec. \ref{subsec:visualization} shows the visualization analysis and Sec. \ref{subsec:SOTA} presents the comparisons with the state-of-the-art approaches for both DG and UDA.

\subsection{Datasets and Implementation Details}\label{subsec:dataset}
We test our method on the four classification datasets of multiple domains: Digit-Five, DomainNet \cite{peng2019moment}, PACS, Office-Home \cite{venkateswara2017Deep}. Figure~\ref{fig:datasets} shows some samples and please see more details in \textbf{Supplementary}. We follow prior works \cite{li2017deeper,cvpr19JiGen,li2019episodic} to use the leave-one-domain-out protocol. Following \cite{peng2019moment}, we build the backbone using three Conv-layers and two FC-layers for Digit-Five. We use ResNet18 \cite{he2016deep} as the backbone for mini-DomainNet, PACS, and Office-Home. More training details are presented in \textbf{Supplementary}.



\subsection{Ablation Study}\label{subsec:ablation}



\begin{table}[htbp]
  \centering
  \footnotesize
  \vspace{-4mm}
  \caption{Effectiveness of our FAR. Note that the \textit{italics} denotes the left-out target domain.}
  \setlength{\tabcolsep}{1.6mm}{    \begin{tabular}{l|ccccc|c||cccc|c}
    \hline
    \multicolumn{1}{c|}{\multirow{2}[1]{*}{Method}} & \multicolumn{6}{c||}{Digit-Five}              & \multicolumn{5}{c}{mini-DomainNet} \\
\cline{2-12}          & \textit{mm} & \textit{mt} & \textit{up} & \textit{sv} & \textit{syn} & \textit{Avg} & \textit{clp} & \textit{pnt} & \textit{rel} & \textit{skt} & \textit{Avg} \\
    \hline
    Baseline & 63.37 & 90.50 & 88.71 & 63.54 & 82.44 & 77.71 & 62.86 & 47.94 & 58.73 & 43.02 & 53.14 \\
    Baseline-align & 65.01 & 93.18 & 93.12 & 76.41 & 84.59 & 82.46 & 63.17 & 49.21 & 58.10 & 44.76 & 53.81 \\
    Baseline-att-align & 68.17 & 96.86 & 94.23 & 78.11 & 91.89 & 85.85 & 65.56 & 47.62 & 57.78 & 48.73 & 54.92 \\
    \hline
    FAR   & \textbf{72.27} & \textbf{99.12} & \textbf{98.60} & \textbf{86.52} & \textbf{95.44} & \textbf{90.39} & \textbf{66.51} & \textbf{50.48} & \textbf{61.11} & \textbf{50.00} & \textbf{57.03} \\
    \hline
    \end{tabular}}%
  \label{tab:ablationstudy_1}%
\end{table}%

\noindent\textbf{Effectiveness of our FAR.}
We compare several schemes. \textbf{\emph{Baseline}}: combines all source domains for training with standard supervised learning. \textbf{\emph{Baseline-align}}: a naive model that builds upon \emph{Baseline}, where we add the moment alignment  constraint on the extracted feature $F$. \textbf{\emph{Baseline-att-align}}: we add Spatial$\&$Channel attention (abbr., SCA) on feature $F$ and add the moment alignment constraint on $A$. \textbf{\emph{FAR}}: our final scheme where the proposed FAR is added after \emph{Baseline} (see Figure~\ref{fig:pipeline}). Table \ref{tab:ablationstudy_1} shows the results. We have the following observations:

\noindent\textbf{1)} \emph{Baseline-align} outperforms \emph{Baseline} by \textbf{4.75\%} in accuracy on Digit-Five, which demonstrates the effectiveness of moment alignment for UDA.

\noindent\textbf{2)} \emph{Baseline-att-align} further improves over \emph{Baseline-align} by \textbf{3.39\%} on Digit-Five, and \textbf{1.11\%} on mini-DomainNet. The attentive gating could facilitate the adaptive selection of feature sub-spaces that are suitable for alignment. 

\noindent\textbf{3)} By compensating the task-relevant information by the proposed \emph{restoration}, our final scheme \emph{FAR} has  high generalization and discrimination capability. It significantly outperforms all the baseline schemes. In particular, \emph{FAR} outperforms \emph{Baseline-align} by \textbf{7.93\%} and \textbf{3.22\%} on Digit-Five and mini-DomainNet, respectively. Such large improvements demonstrate that feature alignment is not enough, and the proposed restoration is critical.

\begin{table}[htbp]
  \vspace{-3mm}
  \centering
  \footnotesize
  \caption{Ablation study on the dual ranking entropy loss $\mathcal{L}_{DRE}$.} 
  \setlength{\tabcolsep}{1.6mm}{  
    \begin{tabular}{l|ccccc|c||cccc|c}
    \hline
    \multicolumn{1}{c|}{\multirow{2}[1]{*}{Method}} & \multicolumn{6}{c||}{Digit-Five}              & \multicolumn{5}{c}{mini-DomainNet} \\
\cline{2-12}          & \textit{mm} & \textit{mt} & \textit{up} & \textit{sv} & \textit{syn} & \textit{Avg} & \textit{clp} & \textit{pnt} & \textit{rel} & \textit{skt} & \textit{Avg} \\
    \hline
    Baseline & 63.37 & 90.50 & 88.71 & 63.54 & 82.44 & 77.71 & 62.86 & 47.94 & 58.73 & 43.02 & 53.14 \\
    FAR w/o $\mathcal{L}_{DRE}$ & 69.82 & 98.09 & 98.33 & 84.63 & 93.37 & 88.85 & 63.51 & 50.24 & 59.88 & 49.87 & 55.88 \\
    FAR w/o $\mathcal{L}_{DRE}^+$ & 70.98 & 98.57 & 98.38 & 84.95 & 93.62 & 89.30 & 64.29 & 50.00 & 60.48 & \textbf{50.16} & 56.23 \\
    FAR w/o $\mathcal{L}_{DRE}^-$ & 71.55 & 98.78 & 98.16 & 85.05 & 94.21 & 89.55 & 65.40 & \textbf{51.43} & 59.68 & 50.00 & 56.63 \\
    FAR   & \textbf{72.27} & \textbf{99.12} & \textbf{98.60} & \textbf{86.52} & \textbf{95.44} & \textbf{90.39} & \textbf{66.51} & 50.48 & \textbf{61.11} & 50.00 & \textbf{57.03} \\
    \hline
    \end{tabular}}%
  \label{tab:ablationstudy_loss}%
\end{table}%

\noindent\textbf{Effectiveness of Dual Ranking Entropy Loss.}
We perform ablation study on the proposed dual ranking entropy loss $\mathcal{L}_{DRE}$. Table \ref{tab:ablationstudy_loss} shows that our final scheme \emph{FAR} outperforms the scheme without the dual ranking entropy loss (\ieno, scheme \emph{FAR w/o $\mathcal{L}_{DRE}$}) by \textbf{2.07\%} and \textbf{1.15\%} on Digit-Five and mini-DomainNet, respectively. Besides, both the constraint on the \emph{enhanced} feature $\mathcal{L}_{DRE}^+$ and that on the \emph{contaminated} feature $\mathcal{L}_{DRE}^-$ contribute to the good feature disentanglement. 

\begin{table}[htbp]
  \centering
  \footnotesize
  \vspace{-5mm}
  \caption{Study on different design choices of FAR.}
  \setlength{\tabcolsep}{0.8mm}{
    \begin{tabular}{cc|ccccccc}
    \hline
    Source & Target & Baseline & FAR${_{conv}}$  & FAR${_{Gate_C}}$     & FAR${_{Gate_S}}$     & FAR (ours) & w/o ranking & w/o TS learning \\
    \hline
    \textit{mt, up, sv, syn} & \textit{mm} & 63.37 & 65.66 & 66.73 & 66.98 & \textbf{72.27} & 68.98 & 70.98 \\
    \textit{mm, up, sv, syn} & \textit{mt} & 90.50 & 94.96 & 97.03 & 98.99 & 99.12 & 97.49 & \textbf{99.49} \\
    \textit{mt, mm, sv, syn} & \textit{up} & 88.71 & 95.23 & 98.66 & 98.39 & \textbf{98.60} & 98.39 & 98.39 \\
    \textit{mt, mm, up, syn} & \textit{sv} & 63.54 & 81.32 & 83.41 & 84.41 & \textbf{86.52} & 85.32 & 85.24 \\
    \textit{mt, mm, up, sv} & \textit{syn} & 82.44 & 91.43 & 93.18 & 93.21 & \textbf{95.44} & 93.17 & 93.77 \\
    \hline
    \multicolumn{2}{c|}{\textit{Ave.}} & 77.71 & {85.72} & {87.80} & {88.40} & {\textbf{90.39}} & {88.67} & 89.57 \\
    \hline
    \end{tabular}}%
    \vspace{-5mm}
  \label{tab:designs}%
\end{table}%

\subsection{Design Choices of FAR}\label{subsec:design}

\noindent\textbf{Analysis on Attention Design.}
As described in Sec. \ref{subsec:FAR}, we use spatial and channel attention gates to adaptively select features for alignment, and to conduct feature disentanglement within FR. We study the influence of different designs and show the results in Table \ref{tab:designs}. \textbf{\emph{FAR${_{conv}}$}}: map the feature $F$ using 1$\times$1 convolutional layer followed by non-liner ReLU activation to obtain feature $A$. \textbf{\emph{FAR${_{Gate_C}}$}} and \textbf{\emph{FAR${_{Gate_S}}$}}: use channel attention gate alone or spatial attention gate alone.

We observe that 1) our spatial$\&$channel attention  scheme \emph{FAR} outperforms \emph{FAR${_{conv}}$} by \textbf{4.67\%} on Digit-Five, demonstrating the benefit of attention design; (2) \emph{FAR} outperforms \emph{FAR${_{Gate_S}}$}/\emph{FAR${_{Gate_C}}$} by \textbf{1.99\%/2.59\%} on Digit-Five, demonstrating the complementary and comprehensiveness of the two types of attention in our FAR framework.

\noindent\textbf{Is `Ranking' Necessary for $\mathcal{L}_{DRE}$?}
Here `ranking' in $\mathcal{L}_{DRE}$ refers to the operation of comparing the entropy of the predicted class likelihood of features \emph{before} and \emph{after} the feature restoration. To verify the effectiveness this strategy, we compare it with the scheme without ranking, which minimizes the entropy loss of the predicted class distribution of the \emph{enhanced} feature $\mathbf{f^+}$ and maximizes the entropy loss of the predicted class distribution of the \emph{contaminated} feature $\mathbf{f^-}$, \ieno, without comparison with \emph{before}. The results in Table~\ref{tab:designs} reveal that our scheme \emph{FAR} with the ranking idea outperforms the above-mentioned scheme \emph{w/o ranking} by \textbf{1.72\%} on Digit-Five.

\noindent\textbf{Teacher-Student Learning for Classifier.}~\label{ensemble_for_classifier} We compare the case without using the teacher-student (TS) strategy (\ieno, constraint $\mathcal{L}_{consist}$) which is denoted as \emph{w/o TS learning}. As shown in the Table \ref{tab:designs}, this strategy introduces \textbf{0.82\%} gain on average on Digit-Five.

\subsection{Visualization}
\label{subsec:visualization}

\noindent\textbf{Feature Map Visualization.} 
To better understand how the FAR method works, we visualize the intermediate feature maps in different phases. We get each activation map by summarizing the feature maps along channels followed by a spatial $\ell_2$ normalization.

\begin{figure}
  \centerline{\includegraphics[width=0.9\linewidth]{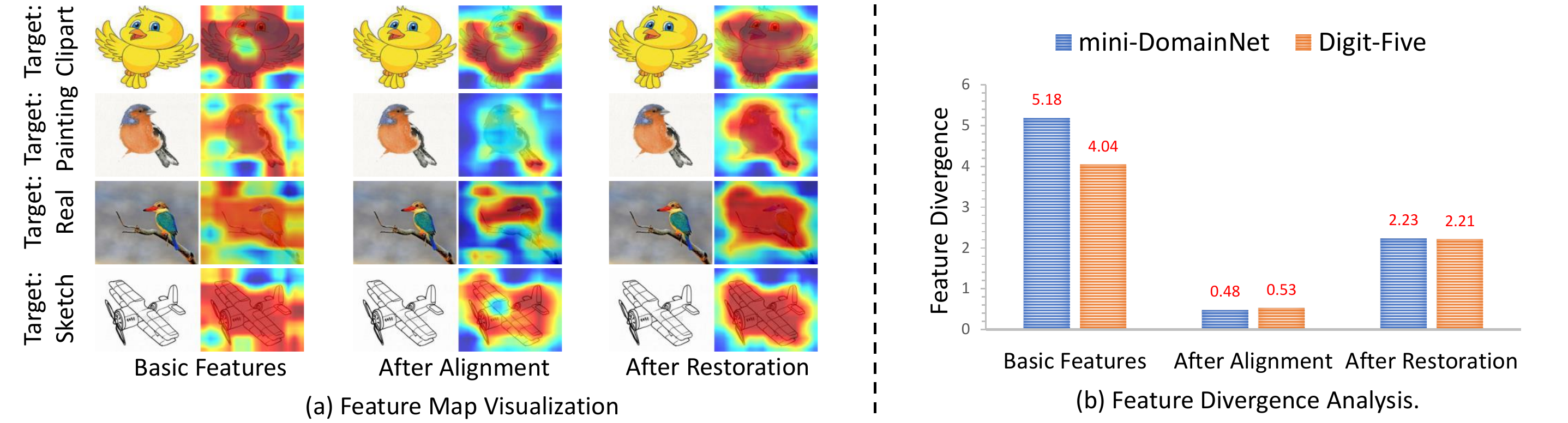}}
    \caption{(a) Activation feature maps in different phases within our FAR framework. They show that the feature alignment (FA) operation may cause the loss of some information that benefits the task, and the feature restoration (FR) could compensate the aligned features with such informative features. (b) Analysis of the feature divergence among different domains in different phases within FAR.}
\label{fig:att}
\vspace{-2mm}
\end{figure}

Figure \ref{fig:att} (a) shows the activation maps of the comprehensive basic feature $F$, aligned feature (\ieno, after FA) $A$, and the final enhanced feature after FR $A+R^+$. We see that the aligned features $A$ typically focus on a more compact region of object to achieve distribution alignment. Since it is task-irrelevant, there is some loss of fine-grained details. Our proposed feature restoration (FR) provides compensation of such task-relevant feature $R^{+}$ and leads to better captured discriminative regions for the final classification. Note that the four images are from different domains of DomainNet. 

\noindent\textbf{Feature Divergence Analysis.} Following \cite{pan2018two,li2016revisiting}, we use the symmetric KL divergence of features between two domains (here we calculate each of the two domains' KL divergence and average them) as the metric to measure their feature divergence. We train the models in \emph{clp,pnt,rel$\rightarrow$skt}/\emph{mm,mt,sv,syn$\rightarrow$up} settings for mini-DomainNet/Digit-Five, and evaluate the feature divergences among all domains (4*250 and 5*200 test samples are randomly selected for mini-DomainNet and Ditgit-Five). We calculate the feature divergence of the basic feature $F$, aligned feature (\ieno, after FA) $A$, and final enhanced feature after FR $A+R^+$.  Figure \ref{fig:att} (b) shows the results.

We observe that the feature divergence (FD) is relatively large for basic feature $F$. The feature alignment operation significantly reduces the FD among domains. Moreover, the FD after feature restoration operation ($A+R^+$) becomes higher than that of $A$, which is because the compensation of some task-relevant features increases the discrimination capability and thus increases the FD.

\begin{figure}
  \centerline{\includegraphics[width=0.8\linewidth]{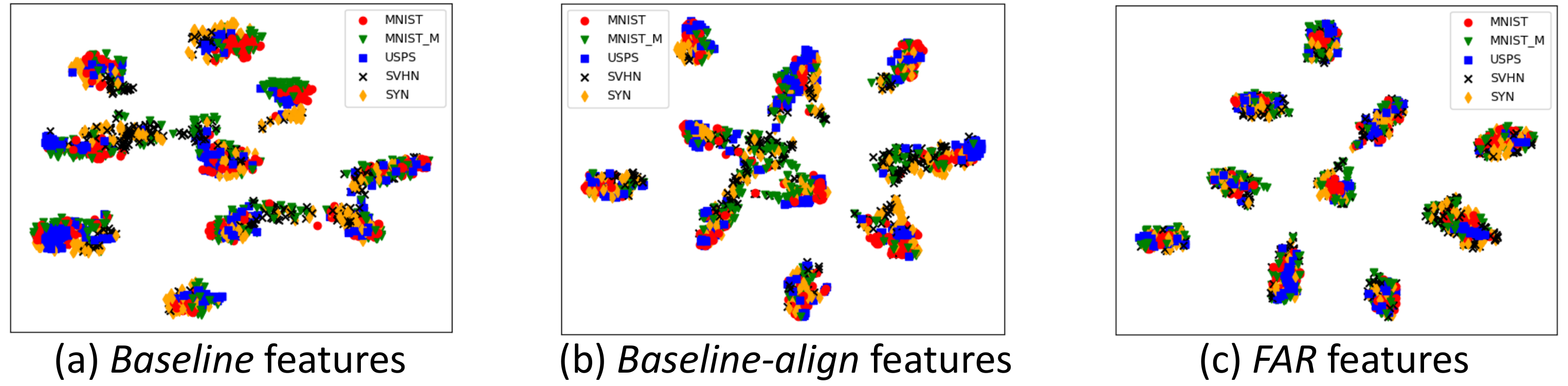}}
    \caption{Visualization of t-SNE distributions. We compare our \emph{FAR} with two baseline schemes.}
\label{fig:tSNE}
\vspace{-3mm}
\end{figure}

\noindent\textbf{Visualization of Feature Distributions.} 
In Figure \ref{fig:tSNE}, we visualize the distribution of the features using t-SNE \cite{maaten2008visualizing} in \emph{mm,mt,sv,syn$\rightarrow$up} setting. We compare the feature distribution of our (c) \emph{FAR} with two baselines (a) \emph{Baseline} and (b) \emph{Baseline-align}. We observe that: 1) feature alignment indeed pulls the different domains closer and makes the distribution compact for each cluster; 2) however, the alignment seems to easily hurt the feature discrimination because it is task-irrelevant, making the different classes closer (indistinguishable); 3) FAR features obtained by our method not only make the domains compact and aligned, but also preserve the clear boundaries of different classes/digits.

\vspace{-2mm}
\subsection{Comparison with State-of-the-Arts}
\label{subsec:SOTA}

\begin{table*}[t]\centering
    \caption{Performance (\%) comparisons with the state-of-the-art approaches for UDA.}
    \vspace{-1mm}
	\captionsetup[subffloat]{justification=centering}
	\subfloat[Comparison results on the Digit-Five. \label{tab:sto_digit5}]
	{
		\tablestyle{3.5pt}{1.1}
            \begin{tabular}{l|ccccc|c}
            \hline
            \multicolumn{1}{c|}{\multirow{2}[1]{*}{Method}} & \multicolumn{6}{c}{Digit-Five} \\
        \cline{2-7}          & \textit{mm} & \textit{mt} & \textit{up} & \textit{sv} & \textit{syn} & \textit{Avg} \\
            \hline
            DAN  \cite{long2015learning} & 63.78 & 96.31 & 94.24 & 62.45 & 85.43 & 80.44 \\
            CORAL  \cite{sun2016return} & 62.53 & 97.21 & 93.45 & 64.40 & 82.77 & 80.07 \\
            DANN  \cite{ganin2016domain} & 71.30 & 97.60 & 92.33 & 63.48 & 85.34 & 82.01 \\
            JAN  \cite{long2017deep} & 65.88 & 97.21 & 95.42 & 75.27 & 86.55 & 84.07 \\
            ADDA  \cite{tzeng2017adversarial} & 71.57 & 97.89 & 92.83 & 75.48 & 86.45 & 84.84 \\
            DCTN  \cite{xu2018deep} & 70.53 & 96.23 & 92.81 & 77.61 & 86.77 & 84.79 \\
            MEDA  \cite{wang2018visual} & 71.31 & 96.47 & 97.01 & 78.45 & 84.62 & 85.60 \\
            MCD  \cite{saito2018maximum} & 72.50 & 96.21 & 95.33 & 78.89 & 87.47 & 86.10 \\
            M3SDA  \cite{peng2019moment} & 69.76 & \textcolor[rgb]{ .357,  .608,  .835}{98.58} & 95.23 & 78.56 & 87.56 & 86.13 \\
            M3SDA-$\beta$  \cite{peng2019moment} & \textcolor[rgb]{ 1,  0,  0}{\textbf{72.82}} & 98.43 & \textcolor[rgb]{ .357,  .608,  .835}{96.14} & \textcolor[rgb]{ .357,  .608,  .835}{81.32} & \textcolor[rgb]{ .357,  .608,  .835}{89.58} & \textcolor[rgb]{ .357,  .608,  .835}{\textbf{87.65}} \\
            \hline
            Baseline & 63.37 & 90.50 & 88.71 & 63.54 & 82.44 & 77.71 \\
            FAR   & \textcolor[rgb]{ .357,  .608,  .835}{72.27} & \textcolor[rgb]{ 1,  0,  0}{\textbf{99.12}} & \textcolor[rgb]{ 1,  0,  0}{\textbf{98.60}} & \textcolor[rgb]{ 1,  0,  0}{\textbf{86.52}} & \textcolor[rgb]{ 1,  0,  0}{\textbf{95.44}} & \textcolor[rgb]{ 1,  0,  0}{\textbf{90.39}} \\
            \hline
            \end{tabular}%
    }
    \hspace{2mm}
	\captionsetup[subfloat]{captionskip=-6pt}
	\subfloat[Comparison results on the mini-DomainNet. \label{tab:sto_mini_domainNet}]
	{
		\tablestyle{4pt}{1.7}
            \begin{tabular}{l|cccc|c}
            \hline
            \multicolumn{1}{c|}{\multirow{2}[1]{*}{Method}} & \multicolumn{5}{c}{mini-DomainNet} \\
        \cline{2-6}          & \textit{clp} & \textit{pnt} & \textit{rel} & \textit{skt} & \textit{Avg} \\
            \hline
            MCD  \cite{saito2018maximum} & 62.91 & 45.77 & 57.57 & 45.88 & 53.03 \\
            DCTN  \cite{xu2018deep} & 62.06 & 48.79 & 58.85 & 48.25 & 54.49 \\
            DANN   \cite{ganin2016domain} & 65.55 & 46.27 & 58.68 & 47.88 & 54.60 \\
            M3SDA  \cite{peng2019moment} & 64.18 & \textcolor[rgb]{ .357,  .608,  .835}{49.05} & 57.70 & \textcolor[rgb]{ .357,  .608,  .835}{49.21} & 55.03 \\
            MME  \cite{saito2019semi} & \textcolor[rgb]{ 1,  0,  0}{\textbf{68.09}} & 47.14 & \textcolor[rgb]{ 1,  0,  0}{\textbf{63.33}} & 43.50 & \textcolor[rgb]{ .357,  .608,  .835}{55.52} \\
            \hline
            Baseline & 62.86 & 47.94 & 58.73 & 43.02 & 53.14 \\
            FAR   & \textcolor[rgb]{ .357,  .608,  .835}{66.51} & \textcolor[rgb]{ 1,  0,  0}{\textbf{50.48}} & \textcolor[rgb]{ .357,  .608,  .835}{61.11} & \textcolor[rgb]{ 1,  0,  0}{\textbf{50.00}} & \textcolor[rgb]{ 1,  0,  0}{\textbf{57.03}} \\
            \hline
            \end{tabular}%
	}
	\label{tab:sto_uda}
	\vspace{-8mm}
\end{table*}

\begin{table}[htbp]
  \centering
  \scriptsize
  \caption{Performance (\%) comparisons with the state-of-the-art approaches for DG.}
  \setlength{\tabcolsep}{2.8mm}{ 
    \begin{tabular}{l|cccc|c||cccc|c}
    \hline
    \multicolumn{1}{c|}{\multirow{2}[1]{*}{Method}} & \multicolumn{5}{c||}{PACS}            & \multicolumn{5}{c}{Office-Home} \\
\cline{2-11}          & \textit{Art}   & \textit{Cartoon}   & \textit{Photo}   & \textit{Sketch}   & \textit{Avg}   & \textit{Art}   & \textit{Clipart}   & \textit{Product}   & \textit{Real}   & \textit{Avg} \\
    \hline
    MMD-AAE  \cite{li2018domain} & 75.2  & 72.7  & 96.0  & 64.2  & 77.0  & 56.5  & 47.3  & 72.1  & 74.8  & 62.7 \\
    CCSA  \cite{motiian2017unified} & \textcolor[rgb]{ .357,  .608,  .835}{80.5} & 76.9  & 93.6  & 66.8  & 79.4  & \textcolor[rgb]{ .357,  .608,  .835}{59.9} & \textcolor[rgb]{ .357,  .608,  .835}{49.9} & 74.1  & 75.7  & 64.9 \\
    JiGen  \cite{carlucci2019domain} & 79.4  & 75.3  & \textcolor[rgb]{ 1,  0,  0}{\textbf{96.2}} & 71.6  & 80.5  & 53.0  & 47.5  & 71.5  & 72.8  & 61.2 \\
    CrossGrad  \cite{shankar2018generalizing} & 79.8  & 76.8  & 96.0  & 70.2  & 80.7  & 58.4  & 49.4  & 73.9  & \textcolor[rgb]{ .357,  .608,  .835}{75.8} & 64.4 \\
    Epi-FCR  \cite{li2019episodic} & \textcolor[rgb]{ 1,  0,  0}{\textbf{82.1}} & \textcolor[rgb]{ .357,  .608,  .835}{77.0} & 93.9  & \textcolor[rgb]{ .357,  .608,  .835}{73.0} & \textcolor[rgb]{ .357,  .608,  .835}{81.5} & -     & -     & -     & -     & - \\
    \hline
    Baseline & 77.0  & 75.9  & \textcolor[rgb]{ .357,  .608,  .835}{96.1} & 69.2  & 79.5  & 58.9  & 49.4  & \textcolor[rgb]{ .357,  .608,  .835}{74.3} & \textcolor[rgb]{ 1,  0,  0}{\textbf{76.2}} & \textcolor[rgb]{ .357,  .608,  .835}{64.7} \\
    FAR   & 79.3  & \textcolor[rgb]{ 1,  0,  0}{\textbf{77.7}} & 95.3  & \textcolor[rgb]{ 1,  0,  0}{\textbf{74.7}} & \textcolor[rgb]{ 1,  0,  0}{\textbf{81.7}} & \textcolor[rgb]{ 1,  0,  0}{\textbf{61.4}} & \textcolor[rgb]{ 1,  0,  0}{\textbf{52.9}} & \textcolor[rgb]{ 1,  0,  0}{\textbf{74.5}} & 75.4  & \textcolor[rgb]{ 1,  0,  0}{\textbf{66.0}} \\
    \hline
    \end{tabular}}%
  \label{tab:sto_dg}%
  \vspace{-4mm}
\end{table}%

\noindent\textbf{Results on Unsupervised Domain Adaptation (UDA).}
Following the leave-one-domain-out protocol \cite{peng2019moment}, the classification accuracy on the target domain test set is reported. Table~(\ref{tab:sto_digit5})(\ref{tab:sto_mini_domainNet}) show the results on the two UDA datasets. We have the following observations. 1) For the overall performance (in the Avg column), FAR achieves the best results on both datasets, outperforming the second-best methods by large margins: 2.74\% on Digit-Five, 1.51\% on mini-DomainNet.
2) On the small Digit-Five dataset, SVHN and SYN are the two relatively difficult domains as shown in Fig.~\ref{fig:datasets}. SVHN contains blurred and cluttered digits while SYN has complex backgrounds. Such large differences of the two domains with other domains make the adaptation harder. Nonetheless, FAR achieves the best accuracy, outperforming M$^3$SDA-$\beta$ \cite{peng2019moment} (2nd best) by 5.20\% on SVHN and 5.86\% on SYN, which further confirms the efficacy of FAR.
3) For the large-scale mini-DomainNet, our FAR also consistently shows clear improvements over the  second-best MME \cite{saito2019semi} on mini-DomainNet. Particularly, in the hardest domain `sketch', FAR achieves the best 50\%, which further validates that feature restoration improves the feature generalization ability, especially for the extreme situation.

\noindent\textbf{Results on Domain Generalization (DG).}
Table~\ref{tab:sto_dg} shows the comparison with the state-of-the-art for DG. We can see that FAR achieves the best results on both PACS and Office-Home with clear margins against all competitors. In particular, FAR is clearly better than the distribution alignment methods, \ieno, CCSA \cite{motiian2017unified} and MMD-AAE \cite{li2018domain}, with 2.3\%/4.7\% improvement on PACS and 1.1\%/3.3\% improvement on Office-Home. 
Besides, compared with the recent self-supervised method JiGen \cite{carlucci2019domain}, data augmentation method CrossGrad \cite{shankar2018generalizing}, and the episodic training method Epi-FCR \cite{li2019episodic}, FAR is clearly superior thanks to the effectiveness of feature alignment and restoration learning. Furthermore, in comparison with the above methods that typically require multiple feature extractors or classifiers, our FAR incurs less computational cost.

\vspace{-2mm}
\section{Conclusion}

In this paper, we study a crucial but overlooked issue that existed in the prior cross-domain generalization/adaptation works, and propose a new concept of Feature Alignment and Restoration (FAR) to help build a unified framework for effective DG and UDA classification. FAR performs feature aligning by moment distributions matching and feature restoration by compensating with task-relevant features to simultaneously enhance cross domain generalization and preserve feature discrimination. For better distilling task-relevant information, we enforce a dual ranking entropy loss constraint to encourage the better disentanglement of task-relevant/-irrelevant feature. 
Extensive experiments on several settings demonstrate the effectiveness of FAR, which achieves the state-of-the-art performance for both domain generalization and unsupervised domain adaptation.

Moreover, we believe that our FAR is a general framework and can be extended to more general and practical fields by equipping with the corresponding task heads, such as image segmentation and object detection. This would be part of our future work.

\clearpage

\appendix
  \renewcommand\thesection{\arabic{section}}

\noindent{\LARGE \textbf{Supplementary}}
\vspace{5mm}

\section{More Details of Feature Alignment and Restoration}

\noindent\textbf{More Details of the Feature Alignment (FA) Phase.} 
To allow the feature $F$ to preserve comprehensive information, instead of on $F$, we enforce the alignment constraints on the adaptively selected feature $A$, which is obtained by gating the comprehensive feature $F$ with the tool of spatial and temporal attention in parallel \cite{hu2018squeeze,woo2018cbam,fu2019dual,liu2019adaptive}. We formulate this as
\begin{equation}
    \begin{aligned}
        A(i,j,:) = S_{i,j} A^{'}(i,j,:), \hspace{0.5cm}
        A^{'}(:,:,k) = a_k F(:,:,k) ,
    \end{aligned}
    \label{eq:seperation}
\end{equation}
where $F(:,:,k) \in \mathbb{R}^{h\times w}$ denotes the $k^{th}$ channel of feature $F$, $k=1,2,\cdots,c$, and $F(i,j,:) \in \mathbb{R}^{c}$ denotes the spatial position of $(i, j)$ of feature $F$, $i=1,2,\cdots,h; j=1,2,\cdots,w$. The channel attention \cite{hu2018squeeze}  response vector ${\mathbf{\textbf{a}}}=[a_1, a_2, \cdots, a_c] \in {\mathbb{R}}^c$ and the spatial attention \cite{woo2018cbam} response map  ${S}\in \mathbb{R}^{h\times w}$ with $S_{i,j}$ denoting the spatial attention value at the position $(i,j)$. They are obtained based on the same input $F$ (\ieno, in parallel instead of in sequential) as
\begin{equation}
    \begin{aligned}
        {\mathbf{\textbf{a}}} &= Gate_{c}(F) = \sigma({\rm W_2^c*}\delta({\rm W_1^c*} pool(F))), \\
        S &= Gate_{s}(F) = \sigma({\rm W_2^s*}\delta({\rm W_1^s*} channel\_pool(F))),
    \end{aligned}
    \label{eq:se}
\end{equation}
where the channel attention gate $Gate_{c}(\cdot)$ consists of a global spatial average pooling layer $pool(\cdot)$ followed by two FC layers with parameters of ${{\rm W_2^c}} \in \mathbb{R}^{(c/r) \times c}$ and ${{\rm W_1^c}} \in \mathbb{R}^{c \times (c/r)}$ which are followed by ReLU activation function $\delta(\cdot)$ and sigmoid activation function $\sigma(\cdot)$, respectively. For the spatial attention gate $Gate_{s}(\cdot)$, it consists of a cross-channel pooling layer, two convolutional layers followed by ReLU and sigmoid activation functions, respectively. To reduce the number of parameters, a dimension reduction ratio $r$ is used and is set to 8 for both channel and spatial gates.

On the attentively selected feature $A$, similar to \cite{peng2019moment}, we add the constraint $\mathcal{L}_{align}$ which aligns the moments of the feature distributions across different domains.
We assume that $A_{s1}$, $A_{s2}$ $, ...,$ $A_{s_{N}}$, $A_{t}$ are the collections of aligned features (originated from i.i.d. samples from the domains $\mathcal{D}_{s1}, \mathcal{D}_{s2}, ..., \mathcal{D}_{s_{N}}, \mathcal{D}_{t}$). The alignment loss $\mathcal{L}_{align}$ in terms of the Moment Distance (MD) between them is defined as
\begin{equation}
    \begin{aligned}
        \mathcal{L}_{align} = MD^2(\mathcal{D}_{s1}, \mathcal{D}_{s2}, ..., \mathcal{D}_{s_{N}}, \mathcal{D}_{t}) &=  \frac{1}{N}\sum_{i=1}^{N} \|{\mu}_i^k - {\mu}_t^k \|_2 + \frac{1}{C_N^2}\sum_{i=1}^{N-1} \sum_{j=i+1}^{N}\| {\mu}_i^k - {\mu}_j^k  \|_2 \\
        &+ \frac{1}{N}\sum_{i=1}^{N} \|{\sigma}_i^k - {\sigma}_t^k \|_2 + \frac{1}{C_N^2}\sum_{i=1}^{N-1} \sum_{j=i+1}^{N}\| {\sigma}_i^k - {\sigma}_j^k  \|_2,
    \end{aligned}
    \label{eq:MD2}
\end{equation}
where $(\mu_i, \sigma_i)$ denotes (mean, variance) of the distributions of aligned features $A_{s_i}$ of source $i$, and $(\mu_t, \sigma_t)$ denotes (mean, variance) of the distribution of aligned features $A_{t}$ of the target domain.

\begin{figure} [b]
  \centerline{\includegraphics[width=1.0\linewidth]{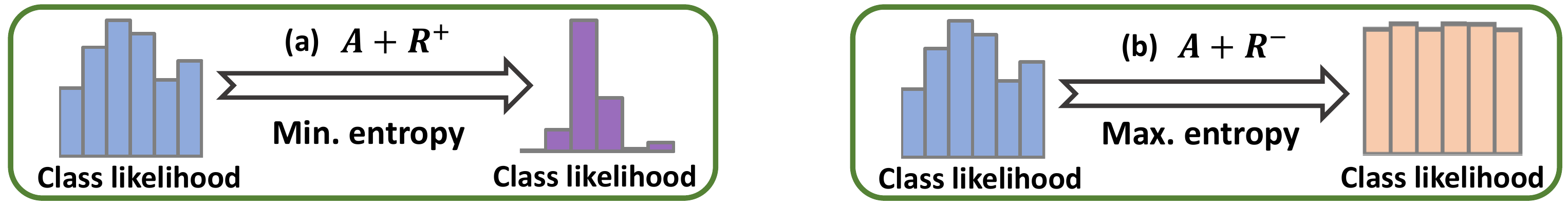}}
    \caption{Illustration of the proposed dual ranking entropy loss constraint. (a) $\mathcal{L}^+_{DRE}$: compensating task-relevant feature benefits classification and then the  predicted class likelihood becomes sharper (smaller entropy), enabling less ambiguous. (b)  $\mathcal{L}^-_{DRE}$: compensating task-irrelevant feature results in larger uncertainty on the predicted class (larger entropy of the predicted class likelihood).}
\label{fig:loss}
\end{figure}

\noindent\textbf{More Details of the Dual Ranking Entropy Loss.} Here we would like to further explain the physical meaning of the proposed dual ranking entropy loss constraint $\mathcal{L}_{DRE}$ within FR phase from a more high-level perspective. As illustrated in Figure \ref{fig:loss}, the main idea is that: after compensating the task-relevant feature $R^+$ to the aligned feature $A$, the feature becomes more discriminative and its predicted class likelihood becomes sharper with smaller entropy, enabling less ambiguous; on the other hand, after compensating the task-irrelevant feature $R^-$ to the aligned feature $A$, the feature should become less discriminative (get larger entropy of the predicted class likelihood).

\section{Training}
Our FAR framework is trained in an end-to-end manner. Here, we introduce more details about the loss for domain generalization and domain adaptation, respectively. The pseudo code of FAR is given in Algorithm 1. 

\noindent\textbf{Domain Generalization.} For DG, the training data is all from the source domains, and the total training loss $\mathcal{L}_{total}$ is a weighted summation of the alignment loss $\mathcal{L}_{align}$, dual ranking entropy loss $\mathcal{L}_{DRE}$, and the basic classification loss (\ieno, cross-entropy loss) $\mathcal{L}_{cls}$:
\begin{equation}
    \begin{aligned}
        \mathcal{L}_{total} = \lambda_{align}\mathcal{L}_{align} +
        \lambda_{DRE}\mathcal{L}_{DRE} + 
        \lambda_{cls}\mathcal{L}_{cls},
    \end{aligned}
    \label{eq:da_loss}
\end{equation}
where $\lambda_{align}, \lambda_{DRE}, \lambda_{cls}$ denote the hyper-parameters for balancing the three losses. Experimentally, we finally set $\lambda_{align} : \lambda_{DRE} : \lambda_{cls} = 0.5:0.1:1$. We follow the common practices to set them. We initially set all of them to 1, and then coarsely determine each one based on the value of each loss and their gradients observed during the training. The principle is to set the balancing weight values to make the weighted loss values/gradients of the three lie in a similar range. Grid search within a small range of the derived value is further employed to get better parameters.  

In addition, as we described in the main manuscripts, to enable fast convergence and learning of the knowledge from the feature $F$ which contains comprehensive information (both domain invariant and domain specific), as a design choice, we assign a classifier on the feature $F$ for each source domain and let them teach the final shared classifier (on the compensated feature $A+R^+$) by enforcing L1-distance consistency $\mathcal{L}_{consist}$ on the predicted probability from the classifiers. Due to the relative smaller loss values in comparison with the other three losses, we experimentally set the balance weight of such sub-loss as $\lambda_{
consist}=100$.   

\noindent\textbf{Domain Adaptation.} For DA, the target domain data without labels is also used, but the loss \emph{for such data} only consists of $\mathcal{L}_{align}$ and $\mathcal{L}_{DRE}$. The loss balance also follows the same rule with DG setting and we finally set them as $\lambda_{align} : \lambda_{DRE} : \lambda_{cls} : \lambda_{consist} = 0.5:0.1:1:100$. 

\noindent\textbf{Pseudo Code of FAR.} The pseudo code of FAR is roughly presented in Algorithm~\ref{alg:far}.

\section{More Details about Datasets and Implementation}
\noindent\textbf{Datasets.} \textit{Digit-5} consists of five different digit recognition datasets: MNIST \cite{lecun1998mnist}, MNIST-M \cite{ganin2015unsupervised}, USPS \cite{hull1994database}, SVHN \cite{netzer2011svhn} and SYN \cite{ganin2015unsupervised}. We follow the same split setting as \cite{peng2019moment} to utilize the dataset. \textit{DomainNet} is a recently introduced benchmark for large-scale multi-source domain adaptation \cite{peng2019moment}, which includes six domains (Clipart, Infograph, Painting, Quickdraw, Real and Sketch) and 600k images with 345 classes. Considering the required considerable computation resources, we takes a subset of DomainNet (\ieno, mini-DomainNet) following \cite{zhou2020domain} for performing experiments.  \textit{PACS}~\cite{li2017deeper} and \textit{Office-Home}~\cite{office_home} are both commonly used DG dataset with four domains, where the former one has seven object categories and the latter one has 65 categories.

\noindent\textbf{Implementation Details.} In all experiments, SGD with momentum is used as the optimizer and the cosine annealing rule~\cite{cosineLR} is adopted for learning rate decay. 

For Digit-5, we build the backbone with three convolution layers and two fully connected layers, following~\cite{peng2019moment}. For each mini-batch, we sample from each domain of 64 images. The model is trained with an initial learning rate of 0.05 for 30 epochs. For mini-DomainNet, we use ResNet18~\cite{he2016deep} as the CNN backbone. We sample 42 images from each domain to form a mini-batch and train the model for 60 epochs with an initial learning rate of 0.005. 

For PACS and Office-Home, similar to \cite{cvpr19JiGen,li2019episodic}, we use ResNet18 as the CNN backbone. We train the model for 40 epochs with an initial learning rate of 0.002. Each mini-batch contains 30 images (10 per source domain).

\begin{algorithm}[t]
   \caption{Feature Alignment and Restoration}
   \label{alg:far}
   \footnotesize
\begin{algorithmic}[1] 
   \STATE {\bf Input:} source domains $\mathcal{D}_{s1}, \mathcal{D}_{s2}...\mathcal{D}_{s_{N}}$, target domain $\mathcal{D}_t$, learning rate $\eta$, batch size $M$, loss balance weights $\lambda_{align} : \lambda_{DRE} : \lambda_{cls} : \lambda_{consist} = 0.5:0.1:1:100$, maximum iterations $N$. The entire network $FAR(\cdot)$ consists of the feature extractor $f_\theta$, feature alignment module $h_\phi$, feature restoration module $r_\psi$, final classifier $c_\omega$, and individual classifier for each source domain $c_{\omega_{i}}$.  For DG, 
  there is no target domain $\mathcal{D}_t$ and the related items can be ignored. 
   \STATE {\bf Output:} predicted class from classifier $c_\omega$.
   \FOR{$n = 1$ {\bf to} $N$}
    \STATE Sample $M$ samples $(x_1, y_1, d_1)$, $\cdots$, $(x_M, y_M, d_M)$ \hfill // \textit{Randomly sample several samples from each domain to form a batch of $M$ samples, where $(x_i, y_i, d_i)$ denotes input sample $x_i$, its class label $y_i$, and domain label $d_i$}
    
     
     \vspace{2mm}
     \STATE  $\mathcal{L}_{total} = \lambda_{align}\mathcal{L}_{align} + \lambda_{DRE}\mathcal{L}_{DRE} + \lambda_{cls}\mathcal{L}_{cls} + \lambda_{consist}\mathcal{L}_{consist}$  \hfill // For each sample in the batch, calculate each loss and the total loss based on the network $FAR(\cdot)$ 
     
     \STATE $(\theta, \phi, \psi, \omega, \omega_{i}) = (\theta, \phi, \psi, \omega, \omega_{i}) - \eta \nabla_{(\theta, \phi, \psi, \omega, \omega_{i})} \lambda_{cls}\mathcal{L}_{cls}$ \hfill // \textit{Use the classification loss $\mathcal{L}_{cls}$ to update the entire network}
     
     \STATE $\phi = \phi - \eta \nabla_{\phi} \lambda_{align}\mathcal{L}_{align}$ \hfill // \textit{Use the alignment loss $\mathcal{L}_{align}$ to update feature alignment module $h_\phi$}     
     
     \STATE $\psi = \psi - \eta \nabla_{\psi} \lambda_{DRE}\mathcal{L}_{DRE}$ \hfill // \textit{Use the proposed dual ranking entropy loss $\mathcal{L}_{DRE}$ to update feature restoration module $r_\psi$}    
     
     \STATE $\omega = \omega - \eta \nabla_{\omega} \lambda_{consist}\mathcal{L}_{consist}$ \hfill // \textit{Use the proposed consist loss $\mathcal{L}_{consist}$ to update feature classifier $c_\omega$ in a teacher-student learning manner}       
     
   \ENDFOR
\end{algorithmic}
\end{algorithm}

\section{Complexity} 
\label{subsec:Complexity}
The model size of our final scheme \emph{FAR} is similar to that of \emph{Baseline} (\egno, ResNet18 as backbone, 11.57 M vs. 11.23 M).


\section{Loss Curves Comparison for the Expert Teaching}

\begin{figure} [t]
  \centerline{\includegraphics[width=0.65\linewidth]{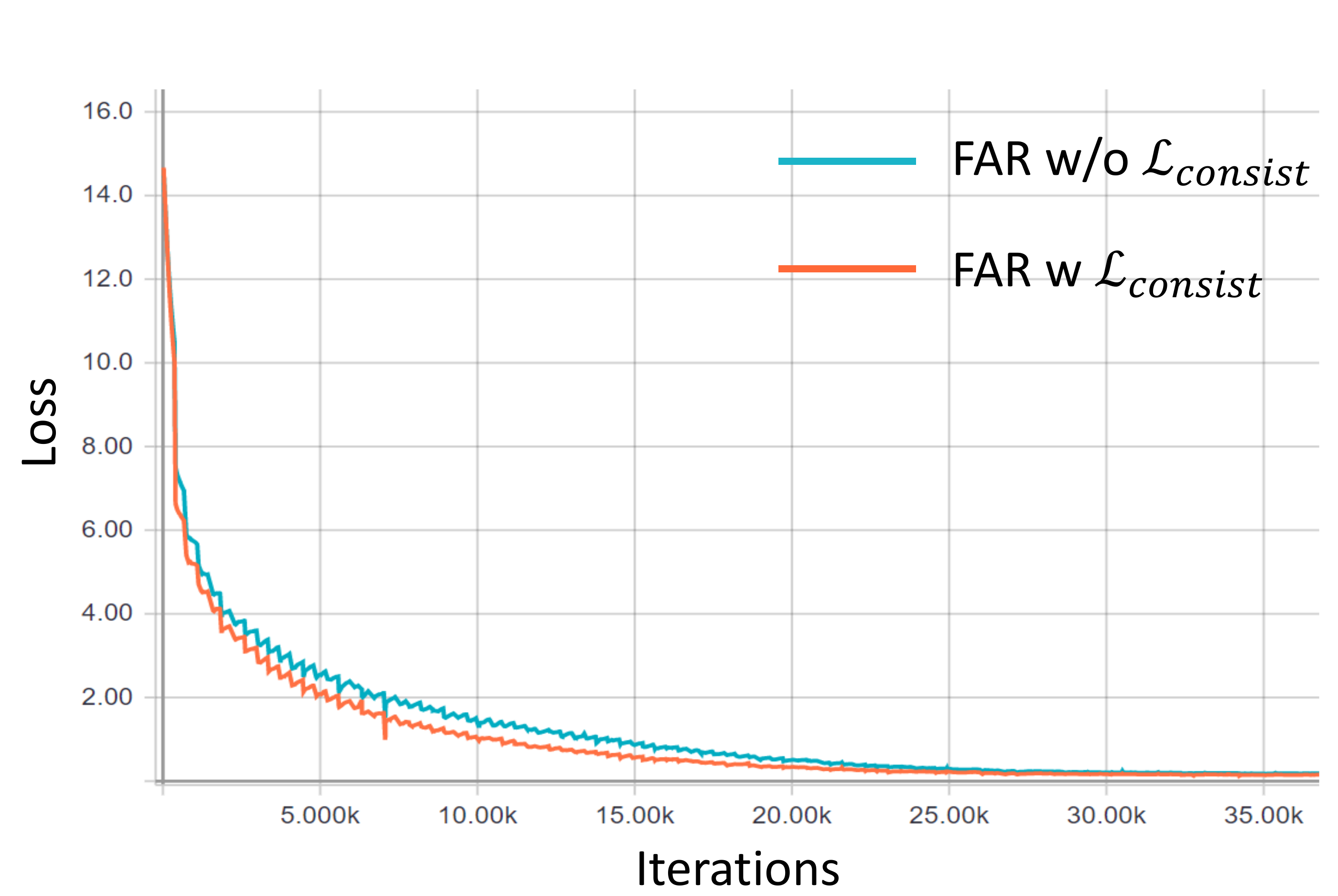}}
    \caption{Loss curves comparison about the expert teaching.}
\label{fig:loss_TS}
\end{figure}

As we have described in the main manuscript, to enable fast convergence and learning of the knowledge from the original feature $F$ which contains comprehensive information (both domain invariant and domain specific), we assign a classifier for each source domain on the feature $F$ and let them teach the final shared classifier by enforcing L1-distance consistency $\mathcal{L}_{consist}$ on the predicted probability from the classifiers. Here we show the loss curves of our FAR without $\mathcal{L}_{consist}$ and our FAR with $\mathcal{L}_{consist}$ in Figure \ref{fig:loss_TS}. We can see that the loss curve of FAR with $\mathcal{L}_{consist}$ converges faster.

{\small
\bibliographystyle{unsrt}
\bibliography{reference}
}

\end{document}